\documentclass[10pt,twocolumn,letterpaper]{article}

\usepackage{cvpr}              

\usepackage[dvipsnames,table,xcdraw]{xcolor}
\usepackage{diagbox}
\usepackage{multirow}
\usepackage{floatrow}
\usepackage{setspace}
\newfloatcommand{capbtabbox}{table}[][\FBwidth]
\usepackage{subcaption}
\usepackage{bbding}
\def\bv #1{\boldsymbol{\rm{#1}}}
\def\etc{{etc}\onedot} 
\def\wrt{w.r.t\onedot}

\newcommand{\eref}[1]{Eq.~(\ref{#1})}
\newcommand{\fref}[1]{Fig.~\ref{#1}}

\newcommand{\tref}[1]{Tab.~\ref{#1}}
\graphicspath{figure}

\definecolor{sred}{rgb}{0.8,0.0,0.0}
\definecolor{sgreen}{rgb}{0,0.8,0}

\usepackage[dvipsnames]{xcolor}

\definecolor{cvprblue}{rgb}{0.21,0.49,0.74}
\usepackage[pagebackref,breaklinks,colorlinks,citecolor=cvprblue]{hyperref}

\def\paperID{5773} 
\def\confName{CVPR}
\def\confYear{2024}

\title{What Do You See in Vehicle? Comprehensive Vision Solution\\ for In-Vehicle Gaze Estimation}

\author{ Yihua Cheng$^1$\quad Yaning Zhu$^2$\quad Zongji Wang$^3$ \quad Hongquan Hao$^4$ \quad Yongwei Liu$^4$\\ Shiqing Cheng$^4$ \quad Xi Wang$^4$ \quad Hyung Jin Chang$^1$\\
University of Birmingham$^1$, \quad Huazhong University of Science and Technology$^2$ \\
 NIST, Chinese Academy of Sciences$^3$, \quad ClamCar$^4$\\
{\tt\small y.cheng.2@bham.ac.uk, h.j.chang@bham.ac.uk}
}

\begin{document}
\maketitle

\begin{abstract}

Driver's eye gaze holds a wealth of cognitive and intentional cues crucial for intelligent vehicles. Despite its significance, research on in-vehicle gaze estimation remains limited due to the scarcity of comprehensive and well-annotated datasets in real driving scenarios.
In this paper, we present three novel elements to advance in-vehicle gaze research.
Firstly, we introduce IVGaze, a pioneering dataset capturing in-vehicle gaze, collected from 125 subjects and covering a large range of gaze and head poses within vehicles. 
In this dataset, we propose a new vision-based solution for in-vehicle gaze collection, introducing a refined gaze target calibration method to tackle annotation challenges.
Second, our research focuses on in-vehicle gaze estimation leveraging the IVGaze. In-vehicle face images often suffer from low resolution, prompting our introduction of a gaze pyramid transformer that leverages transformer-based multilevel features integration. Expanding upon this, we introduce the dual-stream gaze pyramid transformer (GazeDPTR). Employing perspective transformation, we rotate virtual cameras to normalize images, utilizing camera pose to merge normalized and original images for accurate gaze estimation. GazeDPTR shows state-of-the-art performance on the IVGaze dataset.
Thirdly, we explore a novel strategy for gaze zone classification by extending the GazeDPTR. 
A foundational tri-plane and project gaze onto these planes are newly defined. Leveraging both positional features from the projection points and visual attributes from images, we achieve superior performance compared to relying solely on visual features, substantiating the advantage of gaze estimation.
Our project is available at \url{https://yihua.zone/work/ivgaze}.

\end{abstract}

\vspace{-3mm}    
\vspace{-2mm}
\section{Introduction}
\label{sec:intro}
Understanding driver intention and behavior based on driver gaze is in high demand in intelligent vehicles, facilitating diverse applications such as in-vehicle interaction~\cite{app10249011,murali2022intelligent,aftab2019multimodal} and driver monitor systems~\cite{s19112574,s19245540,hayley2021driver}.
Recent advances in vehicle gaze estimation concentrate primarily on gaze zone estimation~\cite{ghosh2021speak2label,jha2018probabilistic,vora2018driver, wang2019continuous}. These approaches define multiple coarse regions, such as side mirrors and windshields, and conduct classification based on face images.
\begin{figure}[t]
    \begin{center}
        \includegraphics[width=0.95\columnwidth]{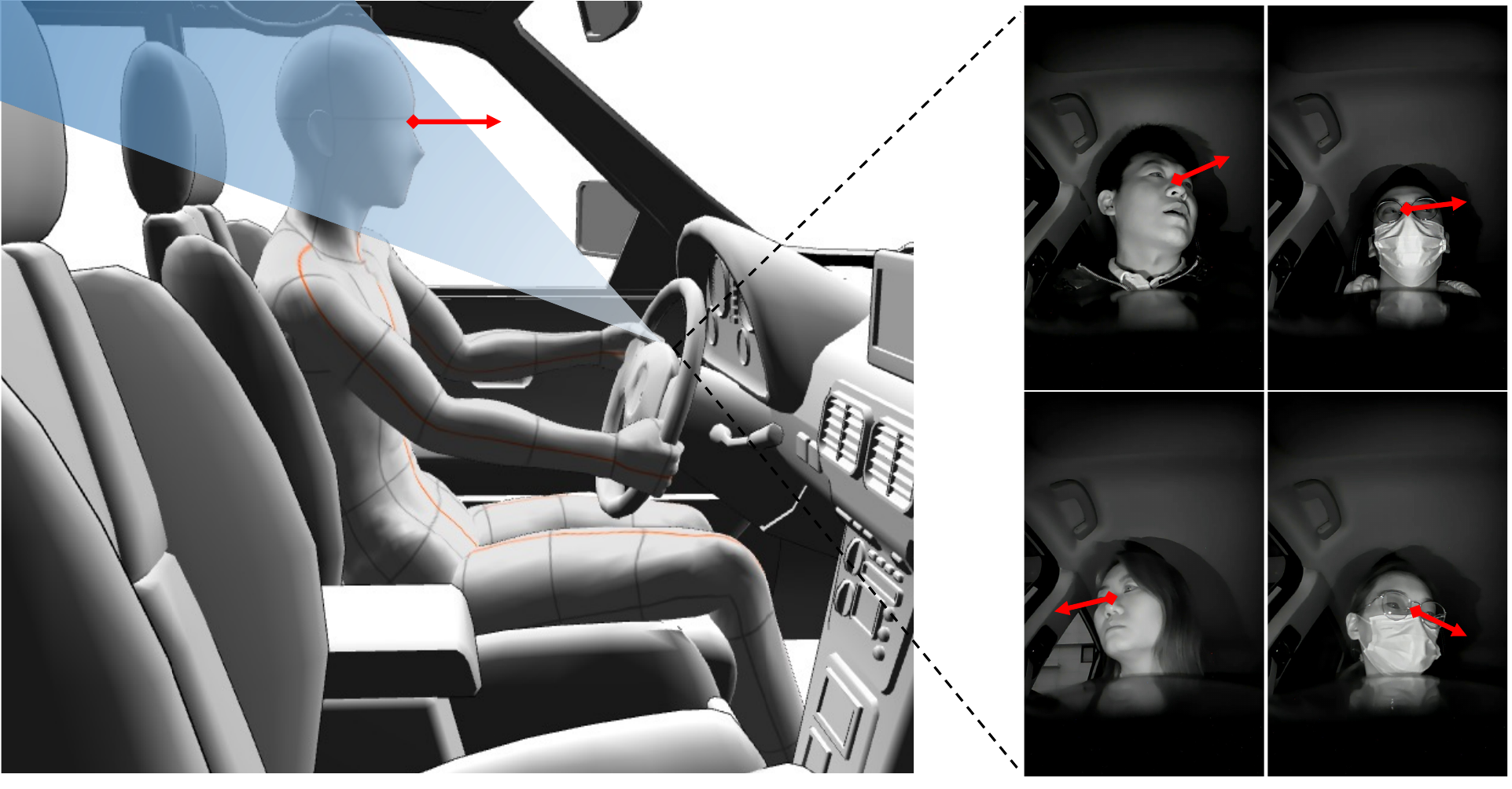}	
    \end{center}
        \vspace{-6mm}
    \caption{In-vehicle gaze estimation illustration. The driver's gaze direction is estimated based on the facial images captured by the camera behind the steering wheels. \vspace{-6mm}}
    \label{fig:calibration}
\end{figure}

Gaze estimation\footnote{Our work focuses on gaze direction estimation. We abbreviate gaze direction as gaze in the rest.} serves as an upstream task of gaze zone estimation and can offer more precise information to understand driver attention.  
However, these methods typically require a large-scale dataset for training.
Although there are numerous gaze datasets, collected in indoor~\cite{Zhang_2017_tpami, Zhang_2020_ECCV} or outdoor~\cite{Kellnhofer_2019_ICCV} environments, their applicability to the vehicle environment is limited due to the different environments and camera settings, resulting in suboptimal performance.
Creating an in-vehicle gaze dataset proves challenging due to the confined and irregular nature of the vehicular environment. Constructing in-vehicle gaze collection systems remains an unsolved issue, as traditional gaze collection systems are impractical for use within vehicles. The absence of in-vehicle gaze datasets acts as a significant barrier to the progress of in-vehicle gaze estimation.

In this paper, we present a comprehensive vision-based in-vehicle gaze estimation research: a novel vision-based gaze collection system for vehicles, offering a first-of-its-kind in-vehicle gaze dataset, a dual-stream gaze pyramid transformer for accurate in-vehicle gaze estimation, and its extension to gaze zone classification, showcasing its effectiveness in enhancing gaze estimation.

First, we introduce IVGaze, an in-vehicle gaze dataset collected from 125 subjects. IVGaze provides a dense distribution of gaze directions, covering a wide range within in-vehicle environments. It contains various conditions, including diverse head poses, eye movements, illumination variations, and the presence of face accessories such as glasses, sunglasses, and masks.
To collect IVGaze, we propose a vision-based gaze collection system.
The system does not require dedicated eye-tracking devices and is easy to reproduce.
We use a single camera for facial appearance capture and paste stickers in vehicles as gaze targets.
However, a significant challenge lies in calibrating the 3D position of gaze targets. This challenge arises because gaze targets are out of the camera's field of view (out-of-FoV). To address this issue, we present a refined gaze target calibration method, which leverages an auxiliary camera to capture gaze targets and a transparent chessboard for calibration.

Secondly, we explore in-vehicle gaze estimation using the IVGaze.
Face images often suffer from low resolution due to the inherent limitations of cameras in vehicles. 
We propose a gaze pyramid transformer that utilizes a transformer to integrate multilevel features.
Expanding upon this, we propose a dual-stream gaze pyramid transformer (GazeDPTR).
We rotate virtual cameras via perspective transformation to normalize images, and leverage camera pose to merge normalized and original images. GazeDPTR shows state-of-the-art results on IVGaze.

Thirdly, we extend GazeDPTR for the downstream gaze zone classification task.
It is challenging to compute the intersection of gaze and the vehicle.
We define a foundational tri-plane and project gaze to the tri-plane. 
We extract positional features from intersection points and predict gaze zones using both positional features and visual features from face images.
Our experiment demonstrates that the gaze zone classification can be further enhanced by positional features, showing the advantage of gaze estimation.


	

 
    
 

\begin{table}[t]
	\caption{\label{tab:DatabseComparison}  The comparison of gaze estimation datasets. Gaze annotation is challenging in the vehicle environment, which results in existing in-vehicle datasets only providing gaze zone annotation. Our work addresses this issue and contributes the first in-vehicle gaze dataset containing gaze annotation and natural face images.} 
	\setlength\tabcolsep{7.5pt}
	\renewcommand\arraystretch{1.0}
	\centering
        \small
	\vspace{-2mm}
	\begin{tabular}{lcccc}
		\toprule[0.4mm]
		\multirow{2}{*}{\textbf{Datasets}}&\multirow{2}{*}{\textbf{Env.}} & \multirow{2}{*}{\# \textbf{Sub.}} & \multicolumn{2}{c}{Annotation}  \\ 
        \cline{4-5}
        &&&Gaze&Zone\\
        \hline
        Gaze360~\cite{Kellnhofer_2019_ICCV}&Outdoor&238 & \color{sgreen}{\Checkmark}& -\\
        \hline
        
        MPIIGaze~\cite{Zhang_2017_tpami}&\multirow{4}{*}{Indoor}  & 15 &\color{sgreen}{\Checkmark} & -\\
		EyeDiap~\cite{Mora_2014_ETRA} & &16& \color{sgreen}{\Checkmark} &-\\
		EVE~\cite{park_2020_eccv}&&54& \color{sgreen}{\Checkmark}& -\\
		ETH-XGaze~\cite{Zhang_2020_ECCV}&&110 & \color{sgreen}{\Checkmark} & -\\
        \hline

		Vora \etal~\cite{vora2018driver} &\multirow{6}{*}{Vehicle} &10  & \color{sred}{\XSolidBrush} & \color{sgreen}{\Checkmark}
		
		\\ 
		
		Jha \& Busso~\cite{jha2018probabilistic}& & 16  &\color{sred}{\XSolidBrush}  &  \color{sgreen}{\Checkmark}
		
		\\ 
		
		Wang \etal~\cite{wang2019continuous}&  & 3     &\color{sred}{\XSolidBrush} &  \color{sgreen}{\Checkmark}
		
		\\ 
		
		Rangesh \etal~\cite{rangesh_2020_IV}&   & 13  &\color{sred}{\XSolidBrush} &  \color{sgreen}{\Checkmark}
		
		\\ 
		
		Ghosh \etal~\cite{ghosh2021speak2label}& & 338  &\color{sred}{\XSolidBrush}  &  \color{sgreen}{\Checkmark}
		\\ 
		\rowcolor[rgb]{0.902,0.902,0.902} \textbf{IVGaze (Ours)} & & 125 &\color{sgreen}{\Checkmark} & \color{sgreen}{\Checkmark}
		\\
		\bottomrule[0.4mm]
	\end{tabular}
        \vspace{-4mm}
        
\end{table}

\section{Related Works}
\label{sec:ralated}

\subsection{Gaze Data Collection}
The human gaze is inherently implicit and poses a challenge for objective measurement, making gaze annotation complex in gaze data collection. 
Some methods capture the human gaze through intrusive devices such as eye-tracking glasses~\cite{Fischer_2018_ECCV,kasahara2022look}.
However, the eye-tracking glasses have a notable impact on the quality of the captured facial images.
Fischer \etal~\cite{Fischer_2018_ECCV} attempted to mitigate this impact using GAN, but there are also some artifacts in the resulting face images.
Vision-based gaze collection systems typically define gaze as direction vectors originating from facial centers towards specific gaze targets~\cite{Zhang_2017_tpami, Zhang_2020_ECCV, park_2020_eccv}.
However, these gaze targets often remain outside the camera's field of view. This out-of-field view challenge complicates the calibration of gaze target positions.
Zhang \etal~\cite{Zhang_2017_tpami, Zhang_2020_ECCV} sets screen points as gaze targets and uses a mirror to calibrate the screen plane.
Kellnhofer \etal~\cite{Kellnhofer_2019_ICCV} uses $360^\circ$ panoramic cameras to capture gaze targets and human faces simultaneously where the panoramic camera is pre-calibrated. However, these strategies are not applicable to vehicles.

In-vehicle gaze dataset usually defines different region such as windshield and left/right mirror in vehicles and perform gaze zone classification~\cite{yuan2022self,ghosh2021speak2label, choi2016real, fridman2016driver, jha2018probabilistic, lee2011real, tawari2014driver}.
Kasahara \etal~\cite{kasahara2022look} collects an in-vehicle gaze dataset but subjects are required to wear eye-tracking glasses, which means the dataset is not applicable in the real world.
Our gaze collection system does not require dedicated devices and produces natural face images.

\begin{figure*}[t]
	\begin{center}
		\includegraphics[width=\columnwidth]{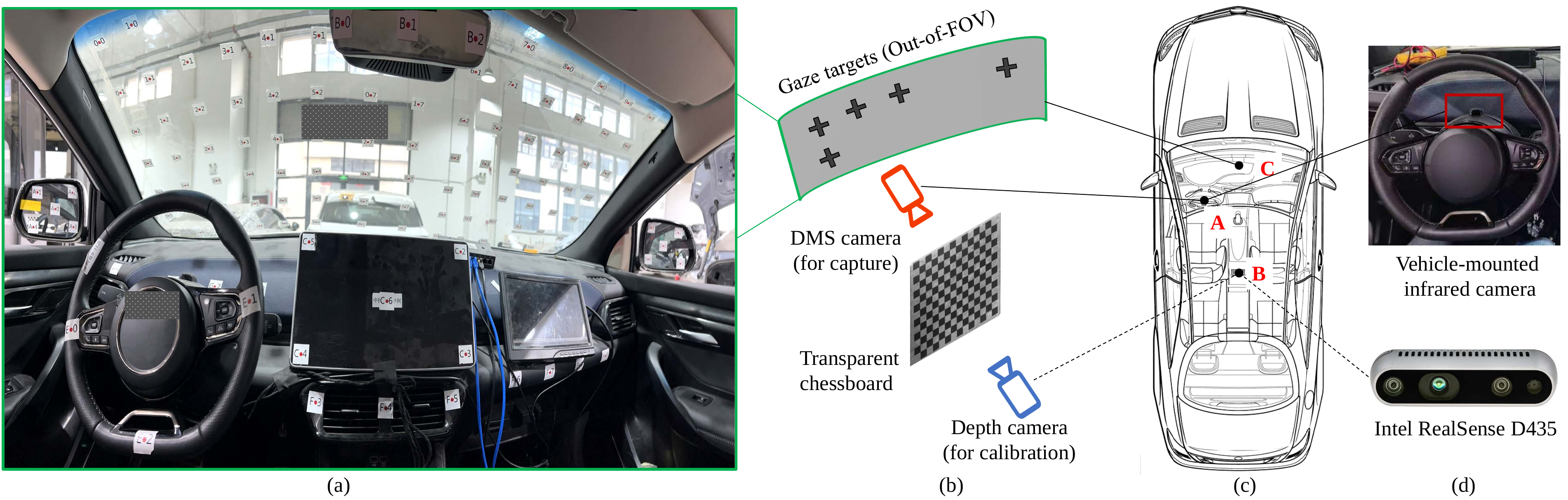}	
	\end{center}
    \vspace{-6mm}
	\caption{
We construct a vision-based in-vehicle gaze collection system comprising a DMS camera, a depth camera, and strategically placed gaze targets, as depicted in (c). The DMS camera is positioned behind the steering wheel to capture drivers' facial appearances, while the gaze targets, positioned beyond the DMS camera's field-of-view (FoV), such as on the windshield, remain unobserved. The depth camera, utilized for calibration purposes, is temporarily installed for capturing gaze target positions in 3D with respect to its own coordinates, and it is removed during data collection. To facilitate the calibration of the depth camera's pose relative to the DMS camera, we propose employing a \textit{transparent chessboard}, which is placed between the two cameras.
	\vspace{-5mm}
        }
	\label{fig:calibration}
\end{figure*}

\subsection{Appearance-based Gaze Estimation}
Appearance-based gaze estimation directly learns mapping function from facial appearance to human gaze~\cite{Cheng_2021_arxiv}.
Conventional gaze estimation methods extract eye features from eye images~\cite{Zhang_2015_CVPR,Cheng_2018_ECCV,Park_2018_ECCV,Cheng_2020_tip}. They concatenate the head pose vector with eye features for gaze estimation.
Recent methods directly learn gaze from face images~\cite{Wang_2022_CVPR, Zhang_2022_CVPR,Cai_2023_CVPR,Bao_2022_CVPR,cheng_2022_aaai}. 
Face images provide both eye region and head pose information, resulting in superior performance compared to methods relying solely on eye images. 
However, the subtlety of the human eye in face images poses a challenge as it can be easily overlooked by the network.
Zhang \etal~\cite{Zhang_2017_CVPRW} utilize a learnable attention map to guide the network's focus specifically on the eye.
Cheng \etal~\cite{cheng_2022_aaai} crop eye images from face images and construct a cascade network to leverage them.
In recent developments, transformers have showcased remarkable capabilities in gaze estimation.
Cheng \etal~\cite{cheng2022icpr} demonstrate that transformers can achieve state-of-the-art performance across various benchmarks.
They apply transformers to address dual-view gaze estimation and show that dual-view images outperform single-view images~\cite{Cheng_2023_ICCV}.


\section{In-Vehicle Gaze Data Collection System}
In this section, we introduce our vision-based system for in-vehicle gaze data collection, eliminating the need for dedicated eye-tracking devices. 
Our system includes a neat gaze target calibration method that effectively addresses the core challenge in gaze annotation. Leveraging this system, we collect the first in-vehicle gaze dataset from 125 subjects. The dataset spans a diverse range of gaze, facilitating and advancing future research in in-vehicle gaze estimation.

\begin{figure*}[t]
	\subfloat[IVGaze dataset samples]{
		\includegraphics[width=0.35\linewidth]{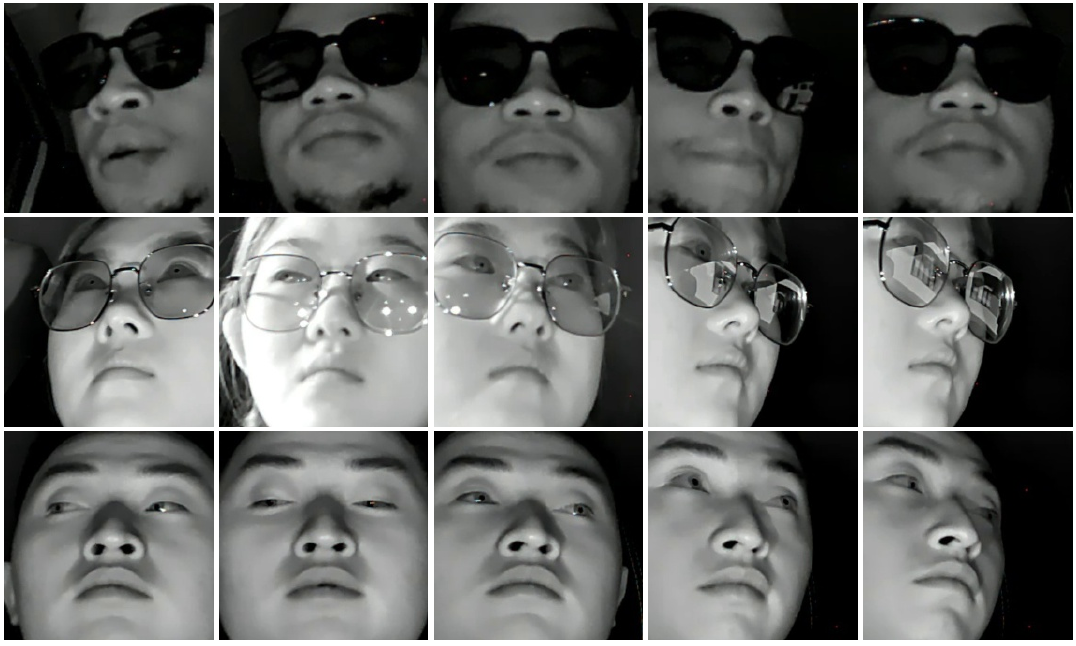}
	}
	\subfloat[Dataset statistics across illumination intensity]{
	    \includegraphics[width=0.4\linewidth]{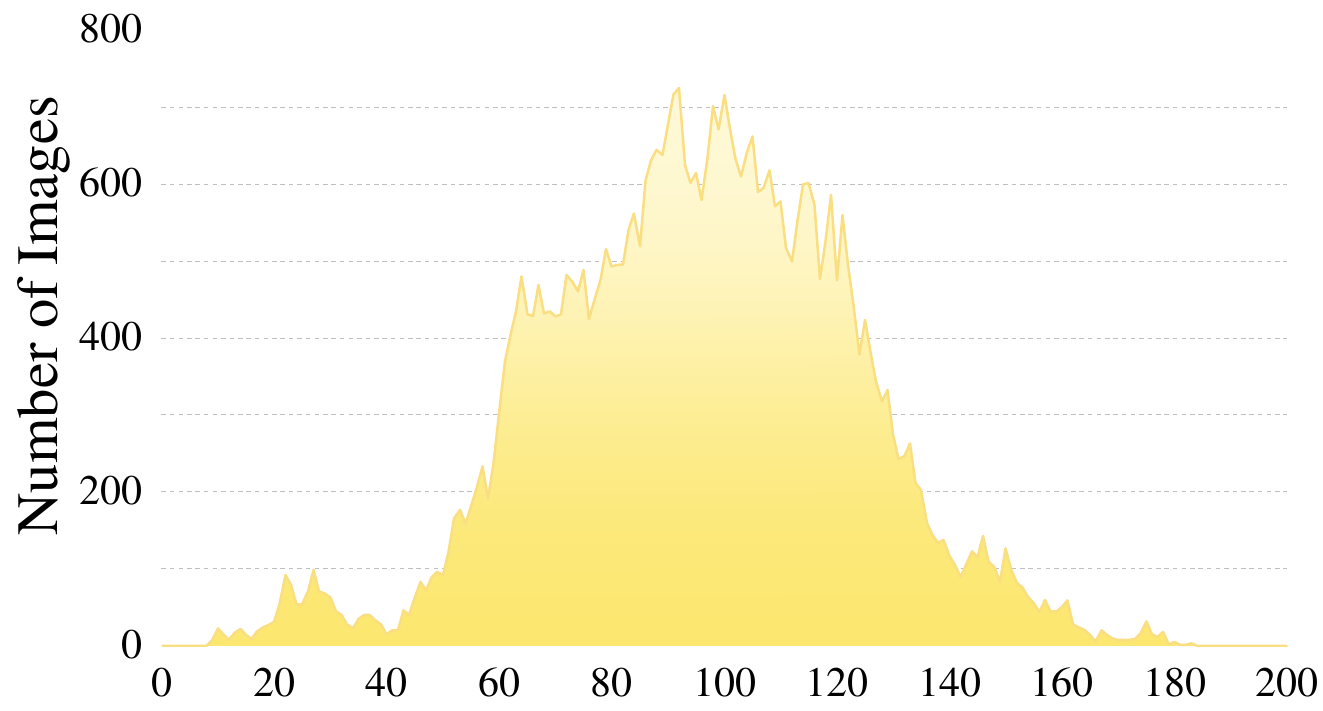}
        }
	\subfloat[Dataset statistics across face accessories]{
		\includegraphics[width=0.27\linewidth]{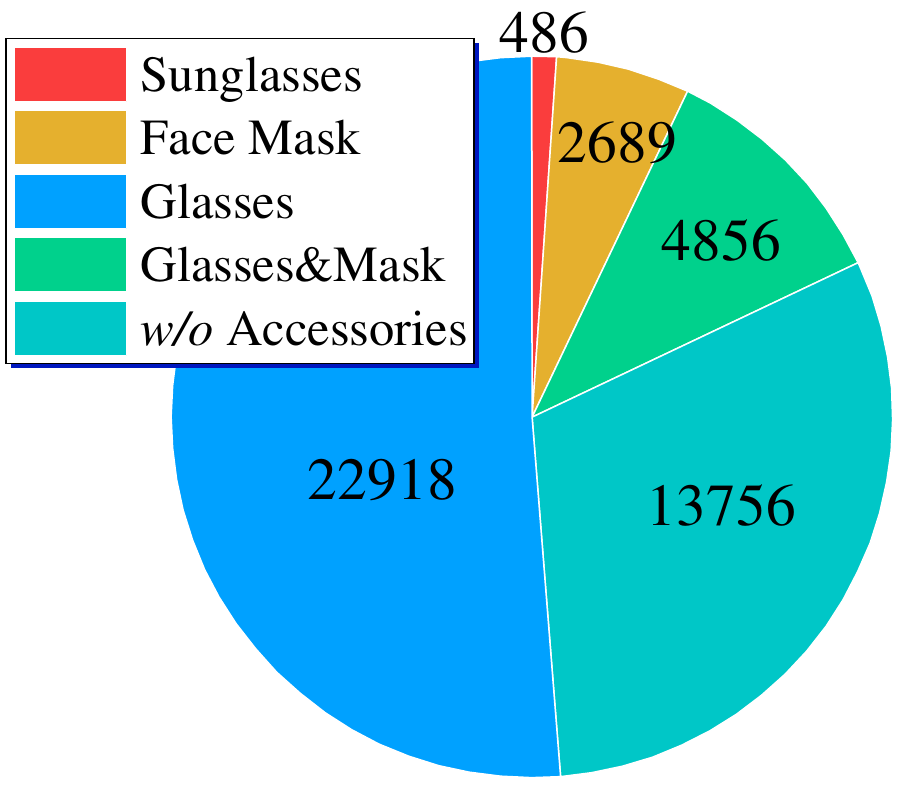}
	}
	\vspace{-2mm}
	\caption{
Our dataset is collected using IR cameras in the vehicle environment.
(a) We present image samples of IVGaze, highlighting the challenges posed by realistic in-vehicle conditions, including cases with sunglasses and reflections in glasses.
(b) We categorize the image count based on their mean pixel value, showing the diversity of illumination conditions.
(c) The image count is analyzed based on face accessories including glasses, sunglasses, and masks.
\vspace{-4mm}
}
	\label{fig:dataset}
\end{figure*}

\subsection{System Setup}
We place gaze targets within the vehicle, and subjects are instructed to look at each designated target during the data collection process. Simultaneously, we record their facial appearance along with corresponding gaze targets.

\textbf{Camera.} Our system uses one camera of a driver monitor system (DMS) for facial appearance capture.
The DMS camera is an infrared camera with a capture resolution of $1280\times800$.
The camera is located behind the steering wheel and directly points at the head region of drivers.

\textbf{Gaze targets.} We mark gaze targets with red points on stickers, and each target is uniquely labeled with a distinct number printed on stickers to aid differentiation.
We strategically position these stickers within various gaze regions within the vehicle, including the windshield, left and right-side mirrors, rear-view mirror, center console, speedometer, handbrake, and \etc. 
These targets cover a large gaze region, satisfying the gaze estimation requirements in vehicles.

\subsection{Collection Procedure}
We meticulously design the collection procedure to minimize error.
During the collection process, participants are instructed to look at specific gaze targets in a predefined sequence. 
They are also required to speak the corresponding number associated with each target to confirm their focus accuracy.
Participants must maintain focus on the target for 3 seconds to facilitate image capture.
We conduct a preliminary check on the captured images and discard any erroneous ones. Once this verification is complete, the current data collection phase concludes, and participants are instructed to shift their gaze to the next gaze targets.

We design three postures for each participant.
Participants should complete the collection process three times with different postures.
The first posture requires participants not to change head pose.
They should preserve the same head pose and only move eyeballs to focus a gaze target.
We aim to collect sufficient eye movement data with this posture.
We do not require participants to focus on the target where participants think it is hard to focus with only eye movement, \eg, the target in the side mirror. 
Participants can rotate their heads to look at these points but they are also required to preserve the new head pose until they cannot look at any targets with the current head pose.
The second posture requires participants not to move their head position but they can rotate their heads to focus a gaze target.
We do not set any constraints in the third posture.
Participants can freely perform head movements in this posture. 
The last two postures bring head pose variation for IVGaze.
We do not use devices such as headrests to constrain participants. Participants only need to preserve the specific posture and focus on targets in a comfortable way.

\subsection{Gaze Annotation}
Vision-based gaze collection systems define gaze as unit direction vectors originating from the face center $\bv{o}$ and extending toward gaze targets $\bv{t}$~\cite{Cheng_2021_arxiv}. Our system follows this framework and the gaze annotation can be computed as $\bv{g} = (\bv{t} - \bv{o})/||\bv{t} - \bv{o}||_2$. The gaze annotation is decomposed into 3D face center estimation and gaze target calibration.

3D face center estimation has been effectively addressed in previous research.
We first detect facial landmarks and fit a 3D morphable face model for 3D facial landmarks~\cite{Zhang_2020_ECCV, lepetit_2009_epnp}. We select the 3D position of the nose as gaze origin  $\bv{o}$.
However, the calibration of gaze target positions presents a challenge because gaze targets are out-of-FoV. It means that the DMS camera cannot capture any images of the gaze targets, complicating the calibration problem.

\subsection{Gaze Target Calibration}
We need the 3D gaze target position $\bv{t}$ for gaze annotation.
However, a key problem is all gaze targets are out-of-FoV. 
Conventional gaze datasets often present gaze targets on a screen and employ a mirror to solve the out-of-FoV problem, where the mirror reflects the content of the screen~\cite{Zhang_2015_CVPR}. However, these methods are not suitable for vehicle scenarios and can lead to a substantial accumulation of errors.

In this section, we propose a neat and efficient gaze target calibration method.
Our basic idea is to employ an auxiliary camera to capture gaze targets.
This allows a straightforward calculation of the 3D gaze target positions \wrt the auxiliary camera coordinate system.
However, it also introduces a new challenge: \textit{how to calibrate extrinsic matrix of the auxiliary cameras \wrt DMS camera}?
Stereo calibration is typically used to compute the pose of two cameras.
However, it requires two captured images sharing corresponding points.
In our system, gaze targets are located behind the DMS camera which means the DMS camera and auxiliary camera should be oriented in opposite directions, making traditional stereo calibration infeasible.

To solve this problem, we propose to use a \textit{transparent chessboard} for the pose calibration.
We set a transparent chessboard between the two cameras and the two cameras respectively capture each side of the chessboard. We first calibrate the two camera poses \wrt~the chessboard coordinate system and then compute the pose between two cameras.
It is worth noting that there are two different chessboard coordinate systems corresponding to each side of the chessboard. We can derive the transformation matrix between the two chessboard coordinate systems where $\bv{R}_\mathrm{chess}=\mathrm{diag}\left(0, 0, -1\right)$ and $\bv{t}_\mathrm{chess} = (0, 0, -\rm{d})$. The $\rm{d}$ is the thickness of the chessboard.

In detail, we use an Intel RealSense D435 depth camera as the auxiliary camera since it can provide accurate 3D positions.
We show the camera layout in \fref{fig:calibration}(b).
We can obtain the transformation matrix ${\bv{R}_\mathrm{dms}, \bv{t}_\mathrm{dms}}$ and ${\bv{R}_\mathrm{depth}, \bv{t}_\mathrm{depth}}$ which can transfer a point from chessboard coordinate systems to camera coordinate systems.
We have 
\begin{equation}
	\bv{R}_\mathrm{rot} = \bv{R}_\mathrm{dms}\bv{R}_\mathrm{chess} \bv{R}_\mathrm{depth}^{-1},
\end{equation}
\begin{equation}
	\bv{t}_\mathrm{rot} = - \bv{R}_\mathrm{dms}\bv{R}_\mathrm{chess} \bv{R}_\mathrm{depth}^{-1}\bv{t}_\mathrm{depth} + \bv{R}_\mathrm{dms}\bv{t}_\mathrm{chess}  + \bv{t}_\mathrm{dms}.
\end{equation}
We use these matrices to convert points from the depth camera coordinate system into the DMS camera coordinate system. Please refer to the supplementary material for details. 


\subsection{In-Vehicle Gaze Dataset}
We collect the first in-vehicle gaze dataset IVGaze which provides dense gaze annotation and natural face images .
We show the dataset samples in \fref{fig:dataset} (a).

\textbf{Dataset Statistics.}
We collect 44,705 images from 125 subjects.
The number of images per subject ranges from 169 to 946.
Most subjects provide around $300$ images per person,~\ie, they repeat the collection procedure three times with three different postures.
40 subjects are female and 85 subjects are male, with ages spanning from 20 to 50 years.

\textbf{Collection Conditions.} 
The dataset was collected between 9 am and 7 pm in outdoor environments, covering a wide range of lighting conditions. 
We also collect images in indoor environments, \ie, a garage.
The dataset statistics across pixel intensity are shown in \fref{fig:dataset} (b).

\textbf{Face Accessories.} We consider two face accessories during the collection: glasses and masks.
For subjects who did not wear glasses, we provided glasses and asked them to repeat the collection wearing glasses.
We provided white and black masks for some subjects and asked them to repeat the collection wearing masks.
We also required a few subjects to wear sunglasses to facilitate future research.
We show the statistics in \fref{fig:dataset} (c).

\begin{figure}[t]
		\includegraphics[width=0.45\linewidth]{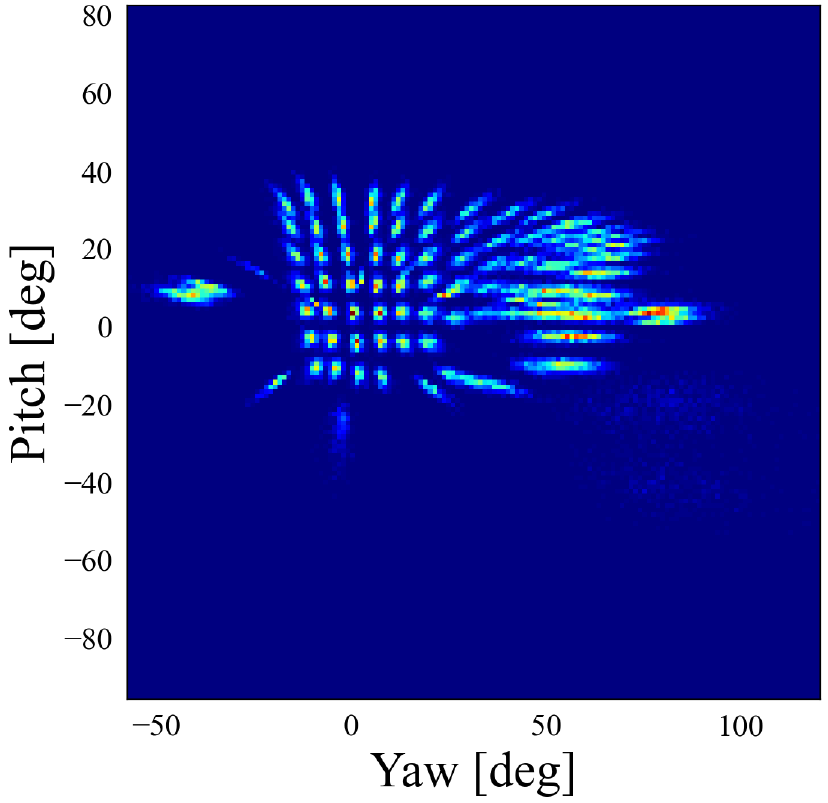}
	    \includegraphics[width=0.46\linewidth]{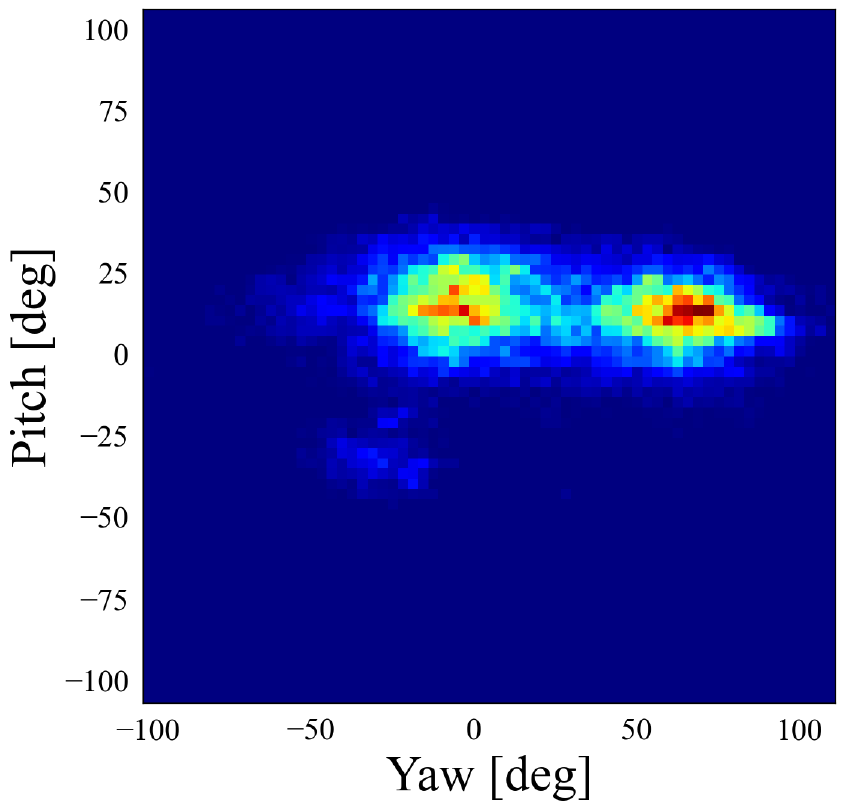}    
     \vspace{-2mm}
	\caption{
We show the distribution of data for gaze (left) and head movements (right). Brighter regions denote higher data density.
    \vspace{-4mm}}
	\label{fig:data_dist}
\end{figure}

\textbf{Data Distribution.}
We plot the angular distribution of gaze and head pose. Note that the data normalization method (introduced in the next section) changes gaze directions and head poses with a rotation matrix. 
\fref{fig:data_dist} shows the distribution in normalization space.
The horizontal gaze is from $-50^\circ$ to $90^\circ$, and the vertical gaze is from $-40^\circ$ to $40^\circ$. It is worth noting that the vehicle environment has limited space, and our dataset already covers almost all regions. Our dataset is collected in the real vehicle environment, and the driver seat in the vehicle is located on the left. Therefore, our dataset has a relatively small distribution on the left. The right figure in \fref{fig:data_dist} shows the head pose distribution. Our dataset contains a large range of head poses in the yaw axis, as subjects need to look at gaze targets from the left-side mirror to the right-side mirror.

\section{In-Vehicle Gaze Estimation}
The in-vehicle environment brings new settings and challenges.
In this section, we systematically explore in-vehicle gaze estimation.
We introduce a revised data preprocessing method and describe a novel gaze estimation network.
We also extend the network for an application and prove the advantage of in-vehicle gaze estimation.

\begin{figure*}[t]
	\begin{center}
		\includegraphics[width=\linewidth]{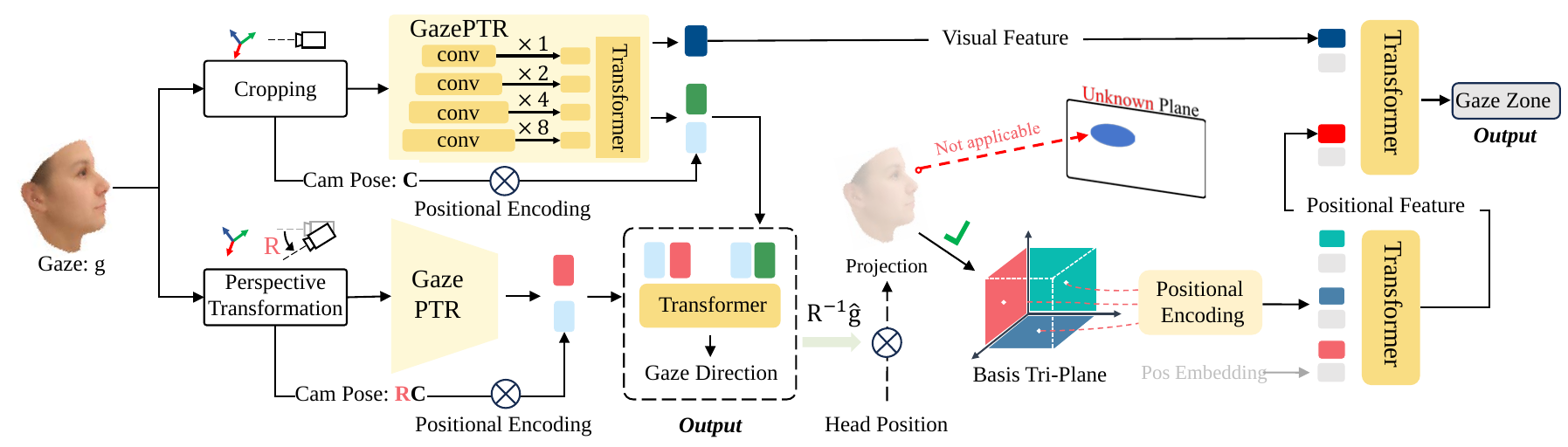}	
	\end{center}
    \vspace{-5mm}
	\caption{
 The GazeDPTR directly crop face for origin images and rotates virtual cameras via perspective transformation for normalized images. It builds a dual-stream network to extract features from the two images based on the  GazePTR for feature extraction which integrates multi-level features via transformers. To further merge the features from two streams, we leverage a transformer where camera pose is used as the positional feature in the transformer. We define the original camera pose as $\bv{C}=diag(1,1,1)$ and the camera pose in normalization space is $\bv{RC}$.  We also extend the network for gaze zone classification. We define a tri-plane and project gaze into them. We extract positional features from three intersection points via a transformer. We also extract visual features from images and predict the gaze zone based on both visual features and positional features. The whole network is trained in an end-to-end manner.
    \vspace{-4mm}}
	\label{fig:network}
\end{figure*}

\subsection{Revisiting Data Normalization in Vehicle}
Data normalization rotates and translates cameras via perspective transformation in face images to reduce head pose variation.
Conventional methods usually define the rotated camera coordinate system $\bv{C}_r$ based on head 
pose~\cite{Sugano_2014_CVPR,Zhang_2018_etra}.
However, these methods cannot bring performance improvement in the vehicle environment.

The $x$-axis of $\bv{C}_r$ is typically defined as the $x$-axis of head coordinate systems.
It means they will rotate images to keep the head straight.
We observed that the designed $x$-axis will produce unstable results, especially in the extreme head pose. 
Therefore, we propose that do not rotate virtual cameras based on the $x$-axis of head coordinate systems.
In detail, we compute the $z$-axis of $\bv{C}_r$ as direction vectors from cameras to face centers.
We use the $x$-axis of the virtual camera, \ie, $(1, 0, 0)$ rather than the head coordinate system.
We compute the $y$-axis as $y=z\times x$ and also recompute the $x$-axis as $x=y\times z$.
Finally, we have a rotation matrix $\bv{R}=[x;y;z]$ and use a scale matrix $S$ to maintain the user-camera distance. 
We warp the image based on matrix $\bv{SR}$.
The human gaze is also changed as $\bv{g}^n = \bv{Rg}^o$, where $\bv{g}^n$ and $\bv{g}^o$ denotes gaze in the normalization and origin space.

\subsection{Gaze Pyramid Transformer}
Gaze estimation methods typically learn gaze from face images.
However, face images captured in vehicles often suffer from low resolution owing to the limited capabilities of the camera. This low resolution can lead conventional methods to overlook subtle details, especially in the eye region.
Some techniques address this challenge by cropping eye images and processing them separately~\cite{Cheng_2020_AAAI}, but these approaches encounter difficulties when faced with extreme head poses, as one of the eye images may not be visible.

We propose a gaze pyramid transformer (GazePTR) in this section.
We build a feature pyramid and input multi-level features into a transformer for gaze estimation.
As shown in \fref{fig:network}, our network uses a convolution network to extract features from facial images. 
We collect feature maps from different levels rather than the last level.
We use $1\times 1$ convolution layers and global average pooling layers to preserve the same feature dimension.
Finally, we input the multi-level feature into a transformer. A learnable token is used to aggregate multi-level features for gaze estimation.

However, the network has the same performance achieved by directly utilizing the last feature map for gaze estimation.
This result can be attributed to the network's tendency to overlook low-level features.
To address this issue, we introduce additional constraints by estimating gaze from each level of features and calculating corresponding loss.
It can be understood that we first extract multi-level features, with each feature being potentially useful. A transformer is used to effectively ensemble these features.

\subsection{Dual-Stream Gaze Pyramid Transformer}
Dual-view images have been demonstrated to outperform single-view images in gaze estimation~\cite{Cheng_2023_ICCV}.
In our work, we rotate cameras in data normalization through perspective transformation.
This inspired us to formulate the dual-camera setting leveraging both original and normalized images.
However, dual-view images improve accuracy since they contain more visual information.
It remains uncertain whether the combination of normalized and original images can provide additional insights beyond what each offers.

To validate our hypothesis, we executed an oracle baseline. We separately train GazePTR on the original and the normalized datasets, selecting the best result from each image pair.
The result reveals a significant performance improvement, validating the inherent advantages of utilizing both images. Please refer to the supplementary for details.

Therefore, we propose a dual-stream gaze pyramid transformer (GazeDPTR). 
Our method leverages both normalized images and original images for gaze estimation. 
We employ GazePTR to extract features from the two images and a transformer is used to integrate the two features for gaze estimation.
Note that, the two features are in different camera coordinate systems and correspond to different gaze. 
To establish the connection, we use the camera pose as the positional information in the transformer.
However, it is impossible the calibrate a virtual camera. 
We define the camera pose of original images as $\bv{C} = diag(1, 1, 1)$ and the camera pose of normalized images is $\bv{RC}$.
We extract the $z$-axis and use positional encoding for camera pose~\cite{Cheng_2023_ICCV}.

\subsection{Strategy for Gaze Zone Classification}

Detecting driver attention is essential in driver monitoring systems. In this section, we introduce a novel strategy to extend GazeDPTR for gaze zone classification. 
GazeDPTR estimates gaze direction from face images.
While the conventional solution involves projecting gaze direction and computing intersections within the vehicle, it is not practical without a 3D vehicle model.
Zhang \etal use a tri-plane for self-supervised loss function~\cite{Zhang_2022_CVPR}.
Our alternative strategy defines a foundational tri-plane and computes the intersection of gaze vectors with this tri-plane.
We extract positional features from three intersection points to facilitate gaze zone classification.

In detail, we define a tri-plane with normal vectors $(1, 0, 0)$, $(0, 1, 0)$, and $(0, 0, 1)$, intersecting at the origin $(0, 0, 0)$. Given the estimated gaze vector $\bv{g}^n$, we convert it back to the original space using $\bv{g}^o = \bv{R}^{-1}\bv{g}^n$. We then project $\bv{g}^o$ onto the tri-plane with the face center $\bv{o}$ and obtain three intersection points. 
To enhance the positional features, we apply positional encoding to these points and input them into a 2-layer transformer. The positional encoding not only increases feature dimensions but also normalizes them within the range of -1 to 1. 
Additionally, we use another learnable token to aggregate visual features from the original images. 
Gaze zone classification is performed using both visual features and positional features.

We train the whole network in an end-to-end manner.
We use L1 loss for gaze estimation and cross-entropy loss for gaze zone classification.
Please refer to the supplementary for implementation details.


\begin{table}[t]
	\setlength\tabcolsep{5pt}
	\renewcommand\arraystretch{1.2}
	\centering
        \small
	\caption{Performance comparison in gaze estimation. Our methods achieve better performance than the compared methods.} 
        \vspace{-2mm}
	\begin{tabular}{lccccc}
		\toprule[1.2pt]
            &\multirow{2}{*}{\shortstack{Angular\\ Error}} & \multicolumn{4}{c}{Average Precision (AP)}\\
            \cline{3-6}
		& &$ \textless 2^\circ$&$ \textless 4^\circ$ &$ \textless 6^\circ$&$ \textless 8^\circ$\\
		\hline
            FullFace~\cite{Zhang_2017_CVPRW}& $13.67^\circ$	& $2.3\%$&	$8.8\%$	&$17.8\%$&$	28\%$\\
            DWG~\cite{ghosh2021speak2label} & $8.82^\circ$	& $6.6\%$  & $21.7\%$ & $38.1\%$&$53.2\%$\\
            Gaze360~\cite{Kellnhofer_2019_ICCV} &$8.15^\circ$	&$9.2\%$	&$27.3\%$	&$44.6\%$	&$58.9\%$\\
            FullFace$^+$~\cite{Zhang_2017_CVPRW}& $7.48^\circ$	&	$14.2\%$&	$31.1\%$&	$46.7\%$&	$63.1\%$\\
            GazeTR~\cite{cheng2022icpr}  & $7.33^\circ$	& $17\%$   & $32.8\%$ & $47.5\%$&$64.7\%$\\
            XGaze~\cite{Zhang_2020_ECCV} & $7.06^\circ$	& $11.7\%$	& $32.7\%$	& $51.5\%$	&$66.7\%$\\
            \hline
            GazePTR & $7.04^\circ$	& $17.6\%$ & $34\%$ & $49.3\%$&$66.7\%$\\
            \rowcolor[rgb]{0.902,0.902,0.902} GazeDPTR & $6.71^\circ$	& $22.1\%$ & $36\%$ & $50.3\%$&$68.4\%$\\
        
		\bottomrule[1.2pt]
	\end{tabular}
 \vspace{-2mm}
	\label{tab:sota}
\end{table}

\section{Experiment}

\textbf{Dataset:} IVGaze contains 44,705 images of 125 subjects.
To perform within-dataset evaluation, we divide the dataset into three subsets based on subjects.
The image numbers of the three subsets are 15,165, 14,674, and 14,866. 
We perform three-fold cross-validation on our dataset.

\noindent \textbf{Evaluation Metric:} We use the angular error as the evaluation metric of gaze estimation as most of the methods~\cite{Cheng_2021_arxiv}, where a smaller value represents a better model. However, we notice the angular error only shows the overall performance while cannot give deep insights. Therefore, we define the average precision (AP) where AP of $\textless k^\circ$ means an estimation is considered correct if the angular error is lower than $k^\circ$.  Regarding the gaze zone classification, average precision is used as a common multi-class classification.

\subsection{Comparison with SOTA Methods}
We first compared our methods with SOTA methods in the in-vehicle gaze estimation task.
We compared our methods with FullFace~\cite{Zhang_2017_CVPRW}, Gaze360~\cite{Kellnhofer_2019_ICCV}, XGaze~\cite{Zhang_2020_ECCV}, DWG~\cite{ghosh2021speak2label} and GazeTR~\cite{cheng2022icpr}.
We replaced the backbone of the FullFace method from AlexNet to ResNet18 for a more convincing comparison. We denote the new method as FullFace$^+$.

The result is shown in \tref{tab:sota}. 
GazePTR and GazeDPTR both show better performance than the compared methods. 
GazePTR and GazeTR have the same backbone and transformer architecture. However, GazePTR outperforms $0.29^\circ$ thanGazeTR since GazePTR uses multi-level feature maps.
GazeDPTR builds a dual-stream network and further brings $0.33^\circ$ improvement than GazePTR. These results show the advantage of our methods.

Interestingly, XGaze has a similar angular error to GazePTR. 
However, it is significantly worse than GazePTR in AP of $\textless 2^\circ$. 
It is because that XGaze uses ResNet50 as the backbone. The deep backbone enhances the feature extraction ability but also easily overlooks the small eye region, making it challenging to achieve precise gaze estimation.


\begin{table}[t]
	\setlength\tabcolsep{4pt}
	\renewcommand\arraystretch{1.2}
	\centering
        \small
	\caption{We show the impacts of different face accessories on performance. 
    The sunglasses bring a significant performance drop.}
        \vspace{-2mm}
	\begin{tabular}{lccccc}
		\toprule[1.2pt]
            \multirow{2}{*}{}& \multicolumn{2}{c}{Glasses} & \multicolumn{2}{c}{Mask}&Sunglasses\\
            \cline{2-6}
		&with & \textit{w/o}&with & \textit{w/o}&with\\
            \hline
            FullFace~\cite{Zhang_2017_CVPRW}&$14.43^\circ$	& $12.40^\circ$  & $15.20^\circ$ &$13.35^\circ$&$21.39^\circ$\\
		DWG~\cite{ghosh2021speak2label} & $9.20^\circ$	& $8.19^\circ$  & $9.43^\circ$ &$8.69^\circ$&$17.43^\circ$\\
            Gaze360\cite{Kellnhofer_2019_ICCV}&$8.30^\circ$	& $7.91^\circ$  & $8.95^\circ$ &$7.99^\circ$ & $17.99^\circ$\\
            FullFace$^+$~\cite{Zhang_2017_CVPRW}&$7.59^\circ$	& $7.30^\circ$  & $8.37^\circ$ &$7.30^\circ$&$16.50^\circ$\\
            XGaze~\cite{Zhang_2020_ECCV} & $7.07^\circ$	& $7.03^\circ$  & $7.80^\circ$ &$6.90^\circ$& $15.15^\circ$\\
            GazeTR~\cite{cheng2022icpr}  & $7.40^\circ$	& $7.22^\circ$  & $8.12^\circ$ &$7.17^\circ$&$17.49^\circ$\\
            \hline
            GazePTR & $7.13^\circ$	& $6.90^\circ$  & $7.78^\circ$ &$6.89^\circ$&$16.54^\circ$\\
            \rowcolor[rgb]{0.902,0.902,0.902} GazeDPTR & $6.77^\circ$	& $6.63^\circ$  & $7.44^\circ$ &$6.57^\circ$&$16.41^\circ$\\
		\bottomrule[1.2pt]
	\end{tabular}
    
	\label{tab:faceacc}
\end{table}

\begin{figure}[t]
		\includegraphics[width=0.49\linewidth]{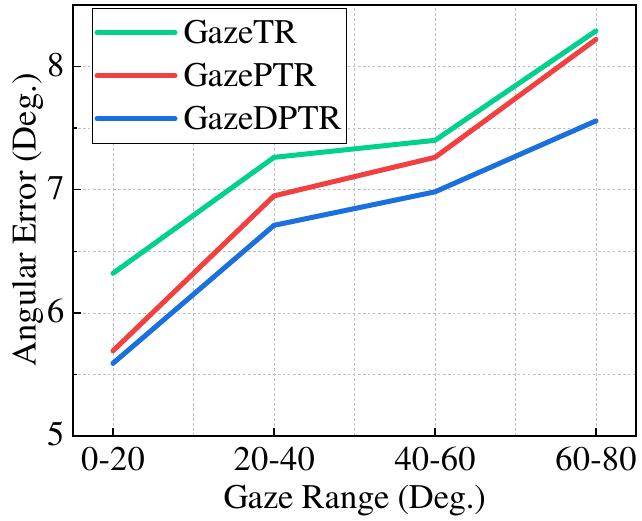}
	    \includegraphics[width=0.49\linewidth]{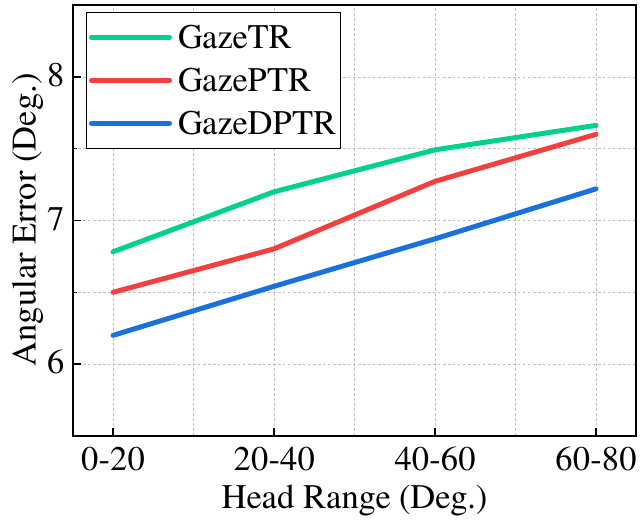}    
	\caption{We compute the angular degree between gaze direction/head orientation with the frontal direction, \ie, (0, 0, -1), and count the mean gaze estimation accuracy in different ranges. Interestingly, GazePTR brings significant performance improvement to GazeTR in the $0^\circ-20^\circ$ gaze range. This is because subjects prefer to move their eyeball rather than their head when the gaze target is located at a close range. GazePTR utilizes multilevel features and can accurately capture eye movement. \vspace{-4mm} }
	\label{fig:range}
\end{figure}

\subsection{The Impact of Face Accessories}
IVGaze provides rich samples of wearing face accessories, including glasses, masks, and sunglasses.
We also evaluate their impacts on the accuracy of gaze estimation and show in \tref{tab:faceacc}.
The result shows that eyeglasses have a large impact if the basic performance is not good. 
However, as the basic performance improves, the performance difference reduces.
This is because most of glasses are transparent, enabling a robust model to extract features effectively.
Sunglasses have a significant impact on performance, with average performance exceeding $15^\circ$.
Masks have a substantial impact on performance, as they obscure facial regions, complicating the extraction of head-related features.

\subsection{Performance Distribution}
We show the performance distribution to provide deep insight.
We first show the performance distribution in different gaze ranges.
We introduce a novel concept, the gaze range of $a^\circ$-$b^\circ$, which is defined as a set of samples satisfying the requirement that the angular degree between the gaze direction and the frontal gaze direction is in the range. 
The result is shown in \fref{fig:range}.
Interestingly, GazePTR and GazeDPTR have better performance in $0^\circ$-$20^\circ$.
This is because subjects prefer to move their eye rather than head to look at objects in this range.
Our methods leverage multi-level features and easily capture eye movement.
We also show the performance distribution in different head ranges. 
Our methods are robust in different head ranges where the maximum performance difference is only $1^\circ$.

\begin{table}[t]
    
	\setlength\tabcolsep{7pt}
	\renewcommand\arraystretch{1.2}
	\centering
        \small
	\caption{We present the performance of GazePTR at each level of the feature. The higher-level feature achieves better performance. Our method integrates features from all levels and achieves the best performance.}
    \vspace{-2mm}
    \arrayrulecolor[rgb]{0.902,0.902,0.902}
	\begin{tabular}{|lccccc|}
        \hline
		& $1st$ & $2nd$ & $3rd$ & $4th$& Ours\\
		\rowcolor[rgb]{0.902,0.902,0.902} GazePTR& $10.33^\circ$&	$8.59^\circ$&	$7.61^\circ$&	$7.35^\circ$&	$7.04^\circ$\\
        \hline
	\end{tabular}
    \vspace{-1mm}
	\label{tab:levelf}
\end{table}

\begin{table}[t]
    \setlength\tabcolsep{4pt}
    \renewcommand\arraystretch{1.1}
    \centering
    \small
    \caption{We present the performance of GazePTR separately training on the original images and normalized images. GazeDPTR leverages camera pose to establish a connection between the two images and use both two images for gaze estimation. We also removed the camera pose to perform an ablation study. }
    \vspace{-1mm}

     \begin{tabular}{|ccccc|}  
    \hline
         &\multirow{2}{*}{\shortstack{Original \\images}} & \multirow{2}{*}{\shortstack{Normalized \\images}}  & \multirow{2}{*}{\shortstack{GazeDPTR \\(\textit{w/o} camera pose)}} & \multirow{2}{*}{\shortstack{GazeDPTR}}  \\
         
        &&&  \\
        \rowcolor[rgb]{0.902,0.902,0.902}Acc&$7.44^\circ$&$7.04^\circ$&$7.03^\circ$& $6.71^\circ$\\
  
        \hline
    \end{tabular}
       \vspace{-5mm}
     \label{tab:ab2}
\end{table}

\subsection{Ablation Study}

\noindent\textbf{GazePTR} leverages multi-level features for gaze estimation. 
We first conduct an ablation study to demonstrate the advantage.
We obtain the gaze performance of GazePTR at each level of feature and show it in \tref{tab:levelf}.
The result demonstrates a higher-level feature achieves better performance. 
GazePTR integrates multi-level features and achieves the best performance. The result proves the advantage of our design.

\noindent\textbf{GazeDPTR} leverages both original and normalized images along with their corresponding camera poses. To demonstrate the effectiveness of our design, we perform an ablation study in \tref{tab:ab2}. Training GazePTR separately on original and normalized images yields performances of $7.44^\circ$ and $7.04^\circ$ respectively. GazeDPTR, which incorporates both image types, achieves a performance improvement of $0.34^\circ$. We also experimented by excluding the positional encoding of the camera pose in GazeDPTR. Interestingly, the result is comparable to using only normalized images.

\subsection{Additional Experiments}

\noindent\textbf{Data Normalization.}
We compare our method with the data normalization method~\cite{Zhang_2018_etra} in the left of \tref{tab:ab}.
The previous method cannot work well in vehicle environments while our method brings $0.3^\circ$ performance improvement.

\begin{figure}[t]
		\includegraphics[width=0.95\linewidth]{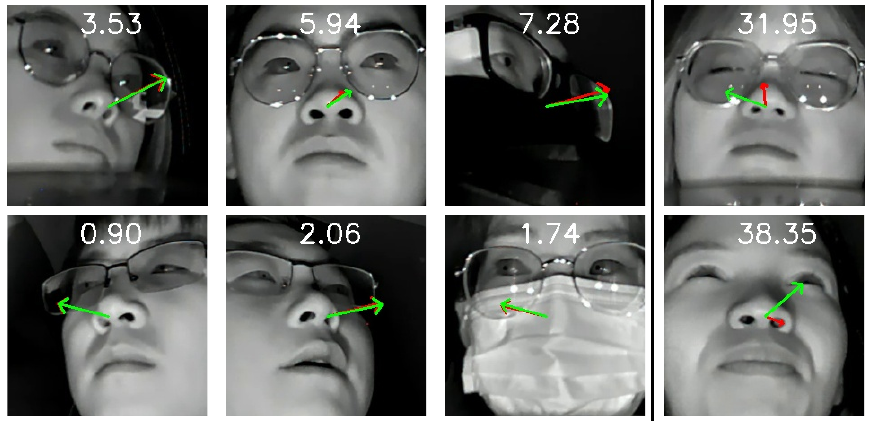}
  \vspace{-2mm}
	\caption{We illustrate {\color{sgreen} GT} with green lines and {\color{red}Prediction} with red lines. The number shows the angular error in degree. The right two figures are failure cases since the eye is hard to capture. \vspace{-3mm}}
	\label{fig:case}
\end{figure}

\noindent\textbf{Case Study.}
We visual gt and our prediction in \fref{fig:case}.

\subsection{Gaze Zone Classification}
We propose a basis tri-plane to acquire positional feature and combine both positional feature and visual features for gaze zone classification.
We respectively evaluate each feature and show the result in the right of \tref{tab:ab}.
The visual feature achieves better performance than the positional feature since the positional feature is obtained from gaze projection. 
The projection increases the model interpretability but decreases the fitting ability.
Our method uses the two features and gets $2.4\%$ performance improvement.
The result demonstrates that the gaze zone classification task could be enhanced by gaze direction cues, highlighting the advantage of our gaze estimation work.

\begin{table}[t]
    \arrayrulecolor[rgb]{0,0,0}
    \setlength\tabcolsep{2pt}
    \renewcommand\arraystretch{1.0}
    \small
    \caption{We evaluate different data normalization methods in the left table. Our method outperforms the previous method~\cite{Zhang_2018_etra}. In the right table, we evaluate different features in the gaze zone classification. Our method uses both positional feature and visual features, achieving $2.4\%$ improvement over visual features.\vspace{-1mm}}
    \hspace{-10mm}
      \begin{subtable}{.4\linewidth}
      \centering
        \begin{tabular}{cc|cc}
            
    \toprule[1.0pt]
        \multicolumn{2}{c|}{Normalization} &\multirow{2}{*}{GazeTR}&\multirow{2}{*}{GazePTR} \\
        \cite{Zhang_2018_etra}&ours&&  \\

                \hline
      		$\times$&$\times$& $7.77^\circ$& $7.44^\circ$\\
                
                \checkmark&& $8.64^\circ$& $8.53^\circ$\\
                \rowcolor[rgb]{0.902,0.902,0.902}&\checkmark&$7.33^\circ$& $7.04^\circ$\\
            \bottomrule[1.0pt]
    	\end{tabular}
    \end{subtable}
    \qquad
    \qquad
    \begin{subtable}{.3\linewidth}
        \centering
	   \begin{tabular}{cc|c}
		  \toprule[1.0pt]
		  \multirow{2}{*}{\shortstack{Visual \\ Feature}}& \multirow{2}{*}{\shortstack{Positional\\ Feature}} &\multirow{2}{*}{AP} \\
  &&\\
		  \hline
		  \checkmark& &	$79.4\%$\\
        &\checkmark&$75.3\%$\\
        \rowcolor[rgb]{0.902,0.902,0.902}\checkmark&\checkmark&$81.8\%$\\
    
		\bottomrule[1.0pt]
	\end{tabular}
    \end{subtable}
    \vspace{-3mm}
     \label{tab:ab}
\end{table}

\vspace{-1mm}
\section{Conclusion}
In this work, we provide systematical research in in-vehicle gaze estimation, including dataset, algorithm, and extensive application.
We first solve the data collection issue in a vehicle where a gaze target calibration method is proposed.
We collect the first in-vehicle gaze dataset containing dense gaze annotation and natural face images.
We further explore the algorithm for in-vehicle gaze estimation.
Our work brings two deep insights 1) multi-level feature is useful to capture eye region information. 2) simultaneously leveraging original images and normalized images could achieve better performance.
We also extend our work for the downstream gaze zone classification task.
We demonstrate that the gaze direction cues could bring performance improvement for gaze zone classification.

 \section{Acknowledgments:}
This work was supported by Institute of Information \& communications Technology Planning \& Evaluation (IITP) grant funded by the Korea government (MSIT) (No.2022-0-00608, Artificial intelligence research about multi-modal interactions for empathetic conversations with humans)
\clearpage
\setcounter{page}{1}
\maketitlesupplementary

Due to the page limitation, we present some details in the supplementary material.
We first describe the details of method and then demonstrate more experiment results.

\section{Methodology}

\subsection{Gaze Target Calibration}
We uses a transparent chessboard for gaze target calibration.
The mathematical deduction is shown in this section.

We set a transparent chessboard between the DMS camera and the depth camera.
The two cameras both capture one side of the chessboard.
Therefore, we can compute the pose matrices of the two cameras \wrt~the chessboard coordinate system.
We denote the pose matrices of the two cameras as $\{\bv{R}_\mathrm{dms}, \bv{t}_\mathrm{dms}\}$ and $\{\bv{R}_\mathrm{depth}, \bv{t}_\mathrm{depth}\}$.
Given a point $\bv{p}_1$ in the chessboard coordinate system, we have
\begin{equation}
	\bv{p}_\mathrm{dms} = \bv{R}_\mathrm{dms}\bv{p}_1  + \bv{t}_\mathrm{dms}.
	\label{equ:dmscamera}
\end{equation}

Similarly, we can compute the 3D position $\bv{p}_\mathrm{depth}$ in the depth camera coordinate system with a given point $\bv{p}_2$.
\begin{equation}
	\bv{p}_\mathrm{depth} = \bv{R}_\mathrm{depth}\bv{p}_2  + \bv{t}_\mathrm{depth}.
	\label{equ:depthcamera}
\end{equation}

Note that, $\bv{p}_1$ and $\bv{p}_2$ represent points in two different chessboard coordinate systems since the two cameras respectively capture each side of the chessboard.
We further derive the rotation matrix and the translation matrix between the two chessboard coordinate systems.
We use $\{\bv{R}_\mathrm{chess}, \bv{t}_\mathrm{chess}\}$ to represent them and have
\begin{equation}
	\bv{p}_1 = \bv{R}_\mathrm{chess} \bv{p}_2 + \bv{t}_\mathrm{chess} 
	\label{equ:chessboard}
\end{equation}

\noindent We have $\bv{R}_\mathrm{chess}=\mathrm{diag}\left(0, 0, -1\right)$ and $\bv{t}_\mathrm{chess} = (0, 0, -\rm{d})$, where $\rm{d}$ is the thickness of the chessboard. Note that, some cameras will capture images in a mirror mode. The rotation matrix should be adjusted based on real setting.

Therefore, given a point $\bv{p}_\mathrm{depth}$ in the depth camera coordinate system, we can obtain the $\bv{p}_\mathrm{dms}$ using \eref{equ:dmscamera},~\eref{equ:depthcamera} and \eref{equ:chessboard}.
We use $\bv{R}_\mathrm{rot}$ and $\bv{t}_\mathrm{rot}$ to represent the rotation and translation matrices between the depth and DMS cameras.
It is easy to derive that
\begin{equation}
	\bv{R}_\mathrm{rot} = \bv{R}_\mathrm{dms}\bv{R}_\mathrm{chess} \bv{R}_\mathrm{depth}^{-1},
\end{equation}
and 
\begin{equation}
	\bv{t}_\mathrm{rot} = - \bv{R}_\mathrm{dms}\bv{R}_\mathrm{chess} \bv{R}_\mathrm{depth}^{-1}\bv{t}_\mathrm{depth} + \bv{R}_\mathrm{dms}\bv{t}_\mathrm{chess}  + \bv{t}_\mathrm{dms}.
\end{equation}

\noindent We can use following equation for the conversion.
\begin{equation}
	\bv{p}_\mathrm{dms} = \bv{R}_\mathrm{rot}\bv{p}_\mathrm{depth}  + \bv{t}_\mathrm{dms}.
	\label{equ:all}
\end{equation}

\subsection{Implementation details of GazeDPTR}
In this paper, we propose a GazeDPTR for gaze estimation. We also extend the GazeDPTR for gaze zone classification.
We train the extended network in an end-to-end manner.

In detail, GazeDPTR contains two GazePTRs for feature extraction from original and normalized images.
GazePTR is modified based on GazeTR~\cite{cheng2022icpr}.
We use ResNet18 to extract multi-level feature maps and obtain 4 different scale feature maps.
Their scales are $64\times56\times56$, $128\times28\times28$, $256\times14\times14$, $512\times7\times7$.
We use $1\times 1$ convolution layers and global average pooling layers to convert them into $128$D features.
We denote these features as $\{f_i\in\mathbb{R}^{128}\}_{i=1,2,3,4}$. We use sup $^n$ and $^o$ to represent the feature is extracted from normalized or original images, \eg $f_1^o$. 
Next, we use a 6-layer transformer to integrate these features for the final feature.
We use one learnable token to aggregate $\{f_i^n\}$ for $f^n_{final}$.
Two learnable tokens are used to aggregate $\{f_i^o\}$ since we need feature $f^o_{final}$ for gaze estimation and visual feature $f_{visual}$ for gaze zone classification.
We use another 6-layer transformer to integrate $f^n_{final}$ and $f^o_{final}$ for $f_{gaze}$.
We add a MLP to estimate gaze directions from $f_{gaze}$. 

\begin{table*}[t]
    \arrayrulecolor[rgb]{0,0,0}
    \setlength\tabcolsep{3pt}
    \renewcommand\arraystretch{1.1}
    \centering
    \small
    \caption{We define nine zones for gaze zone classification and show the average precision ($\%$) on each zone.  }
    \vspace{-2mm}

     \begin{tabular}{|cc|ccccccccc|c|}  
    \hline
         \multirow{2}{*}{\shortstack{Visual \\ Feature}}& \multirow{2}{*}{\shortstack{Positional\\ Feature}} &\multirow{2}{*}{\shortstack{Left-side \\mirror}} &\multirow{2}{*}{\shortstack{Rear-view\\ mirror
}} &\multirow{2}{*}{\shortstack{Right-side\\ mirror}} & \multirow{2}{*}{\shortstack{Central-control\\ screen}}&\multirow{2}{*}{\shortstack{Steering \\wheel}}&\multirow{2}{*}{Handbrake}&\multirow{2}{*}{Dashboard}&\multirow{2}{*}{\shortstack{Left-side \\windshield
}}&\multirow{2}{*}{\shortstack{Right-side \\windshield}}&\multirow{2}{*}{Avg}\\
        &&&&&&&&&&&  \\
        \hline
        \checkmark&&96.5&\underline{47.0}&68.4&\underline{83.4}&\underline{92.5}&79.9&72.5&89.3&69.9&79.4\\ 
        &\checkmark&\underline{97.3}&39.1&\underline{79.5}&63.7&87.6&54.4&45.2&88.9&65.6&75.3\\ 
        \rowcolor[rgb]{0.902,0.902,0.902}\checkmark&\checkmark&96.9&42.1&75.7&77.1&92.0&\underline{86.9}&\underline{77.5}&\underline{89.7}&\underline{75.4}&81.8\\ 
        \hline
    \end{tabular}
       \vspace{-5mm}
     \label{tab:sup-zone}
\end{table*}
We project the estimated gaze into a tri-plane. Note that, we cut off the propagation of gradient in this operation layer since it drops gaze estimation accuracy but cannot improve gaze classification performance.
We use a 2-layer transformer to extract positional feature $f_{pos}$ where a deep transformer will vanish gradients.
We also use a 6-layer transformer to integrate positional features and visual features for  $f_{zone}$.
We add a MLP to predict gaze zone from $f_{zone}$.

Regarding the loss function, we use L1 loss $\mathcal{L}_{gaze}$ for the gaze estimation task.
Our method contains two ground truths $\bv{g}^o$ and $\bv{g}^n$. We define the function  $\mathcal{L}^o_{gaze}(f)$ that means we set a MLP to estimate gaze from feature $f$ and measure the L1 distance between the gaze and $\bv{g}^o$ for loss function. The same for $\mathcal{L}^n_{gaze}(f)$.

we require following feature should be gaze-related including 1) multi-level feature $\{f_i^n\}$ 2) intermediate features $f^n_{final}$ and $f^o_{final}$ 3) gaze feature  $f_{gaze}$. The loss function can be represented as:

\begin{equation}
	\mathcal{L}_{1} = \sum_{i=1}^{4}\sum_{j\in\{o, n\}} \mathcal{L}^j_{gaze}(f^j_i) + \sum_{j\in\{o, n\}} \mathcal{L}^j_{gaze}(f^j_{final}) + \mathcal{L}^n_{gaze}(f_{gaze}) 
	\label{equ:gaze}
\end{equation}

We set cross entropy loss as the loss function for gaze zone classification. We also define the loss $\mathcal{L}_{zone}(f)$ that means we set a MLP to predict gaze zone from $f$ and measure the cross entropy loss. The loss function for gaze zone task is   
\begin{equation}
	\mathcal{L}_{2} = \mathcal{L}_{zone}(f_{pos}) + \mathcal{L}_{zone}(f_{visual}) + \mathcal{L}_{zone}(f_{zone}) 
	\label{equ:zone}
\end{equation}

We optimize the whole network using 
\begin{equation}
	\mathcal{L}_{GazeDPTR} = \mathcal{L}_{1} +  \mathcal{L}_{2}
	\label{equ:zone}
\end{equation}

\begin{table}[t]
    \arrayrulecolor[rgb]{0,0,0}
    \setlength\tabcolsep{4pt}
    \renewcommand\arraystretch{1.1}
    \centering
    \small
    \caption{We define one additional class \textit{None} to account for samples that do not fall within the nine zones. We respectively report the average precision \textit{with} and \textit{w/o} the \textit{None} region. }
    \vspace{-2mm}

    \begin{tabular}{cc|cc}
		  \toprule[1.0pt]
		  \multirow{2}{*}{\shortstack{Visual \\ Feature}}& \multirow{2}{*}{\shortstack{Positional\\ Feature}} &\multirow{2}{*}{\shortstack{AP\\\textit{w/o None} region}} & \multirow{2}{*}{\shortstack{AP\\\textit{with None} region}}\\
  &&\\
		  \hline
		  \checkmark& &	$79.4\%$ &$78.1\%$ \\
        &\checkmark&$75.3\%$ & $75.8\%$\\
        \rowcolor[rgb]{0.902,0.902,0.902}\checkmark&\checkmark&$81.8\%$&$80.0\%$\\
    
		\bottomrule[1.0pt]
	\end{tabular}
       \vspace{-2mm}
     \label{tab:sup-zonecls}
\end{table}

\begin{table}[t]
    \arrayrulecolor[rgb]{0,0,0}
    \setlength\tabcolsep{4pt}
    \renewcommand\arraystretch{1.1}
    \centering
    \small
    \caption{We selected best results from a pair of original and normalized images. The performance shows the selected result significantly outperform each of images.  }
    \vspace{-2mm}

     \begin{tabular}{|cccc|}  
    \hline
         &\multirow{2}{*}{\shortstack{Original \\images}} & \multirow{2}{*}{\shortstack{Normalized \\images}}  & \multirow{2}{*}{Selected}  \\
        &&&  \\
        \rowcolor[rgb]{0.902,0.902,0.902}Acc&$7.44^\circ$&$7.04^\circ$&$5.72^\circ$\\
  
        \hline
    \end{tabular}
       \vspace{-5mm}
     \label{tab:sup-dptr}
\end{table}

\section{Additional Experiments}
\subsection{Setup of Gaze Zone Classification}
Our paper extends gaze estimation for gaze zone classification.
In this section, we provide details about the experimental setup.

During data collection, stickers are placed strategically within the vehicle. Based on the positions of these stickers, we divide the in-vehicle region into nine zones: left-side mirror, right-side mirror, rear-view mirror, steering wheel, left-side windshield, right-side windshield, central-control screen, handbrake, and dashboard. It is important to note that the dashboard encompasses not only the instrument cluster behind the steering wheel but also the air conditioning panel. Additionally, we introduce an extra region \textit{None} to account for points that do not fall within the specified nine zones. The performance of this additional class is not included in the average performance calculation.

The average precision (AP) of each classes is shown in the \fref{tab:sup-zone}. 
GazeDPTR integrates two features and show better average AP.
We also show the average performance with \textit{None} region in \tref{tab:sup-zonecls} for reference.
 
\subsection{Analysis on Normalized and Original Images}
Our work uses both original and normalized images for gaze estimation. 
Our hypothesis is that the conbination of two images can provide additional insights beyond what each offers. 
In this section, we show experimental result to validate our hypothesis.
\begin{figure}[t]
		\includegraphics[width=0.95\linewidth]{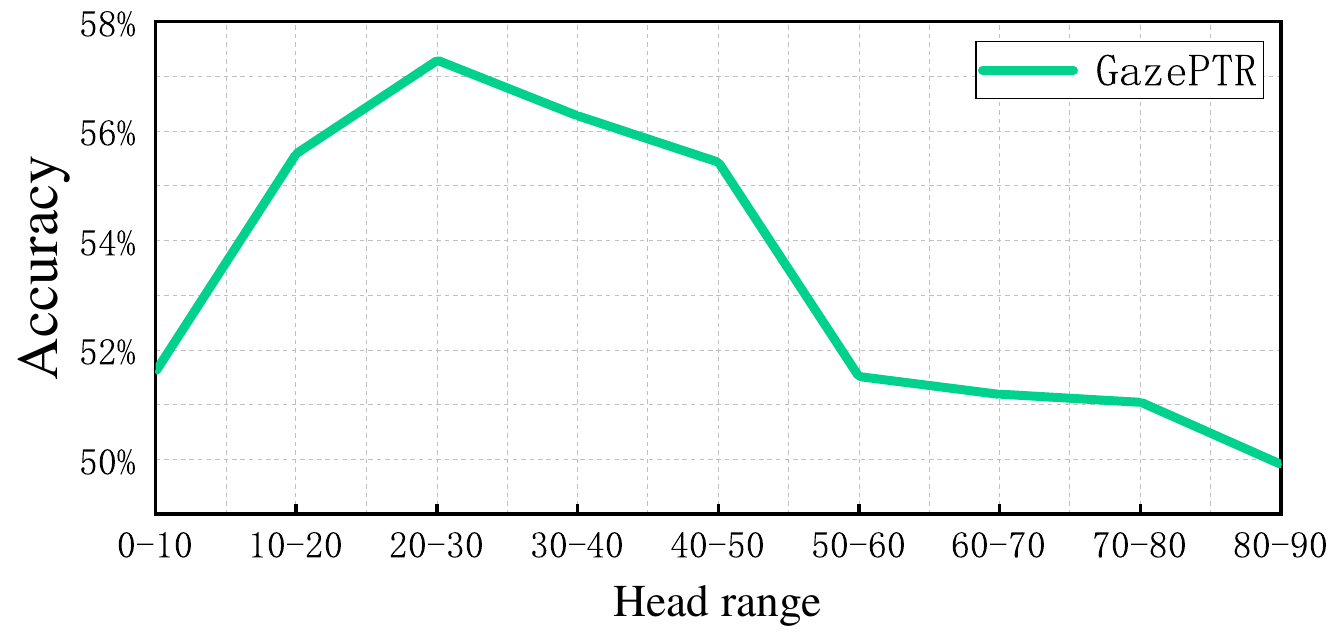} 
	\caption{We count the improvement ratio in each head range. A larger ratio means more samples have performance improvement due to normalization. The result demonstrates that the large head range usually has relatively low improvement ratio.   \vspace{-3mm} }
	\label{fig:sup-range}
\end{figure}
\begin{figure}[t]
	    \includegraphics[width=0.60\linewidth]{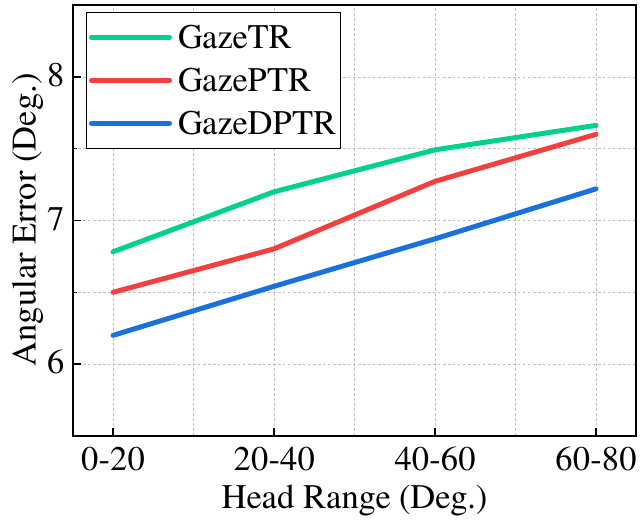}    
	\caption{We count the average angular error in different head ranges. GazePTR estimates gaze from normalized images while GazeDPTR uses both normalized and original images for gaze estimation. It is interesting that GazeDPTR has larger performance improvement in a large head range than GazePTR. Combining with the result in \fref{fig:sup-range}, the reason may be the relatively low improvement ratio in the large head range.\vspace{-3mm} }
	\label{fig:sup-ma-range}
\end{figure}
We initially conducted an oracle baseline by separately training GazePTR on both the original and normalized datasets and selecting the best result from each image pair. The result is shown in \tref{tab:sup-dptr}. The selected performance demonstrated a remarkable improvement, achieving $5.72^\circ$, which significantly surpasses the performances in both the original and normalized images.

To gain a more nuanced understanding of the improvement across different head pose ranges, we calculated the improvement ratio. A particular sample is considered improved if the performance of the normalized image surpasses that of the original image. The results are visualized in \fref{fig:sup-range}, where images that failed in the large head pose range typically exhibit a relatively low improvement ratio. Additionally, the angular error across different head poses is depicted in \fref{fig:sup-ma-range}, underscoring the larger performance improvement in a significant head pose range for GazeDPTR. These findings provide valuable insights into the advantages of our proposed method.

\begin{figure}[t]
	    \includegraphics[width=0.95\linewidth]{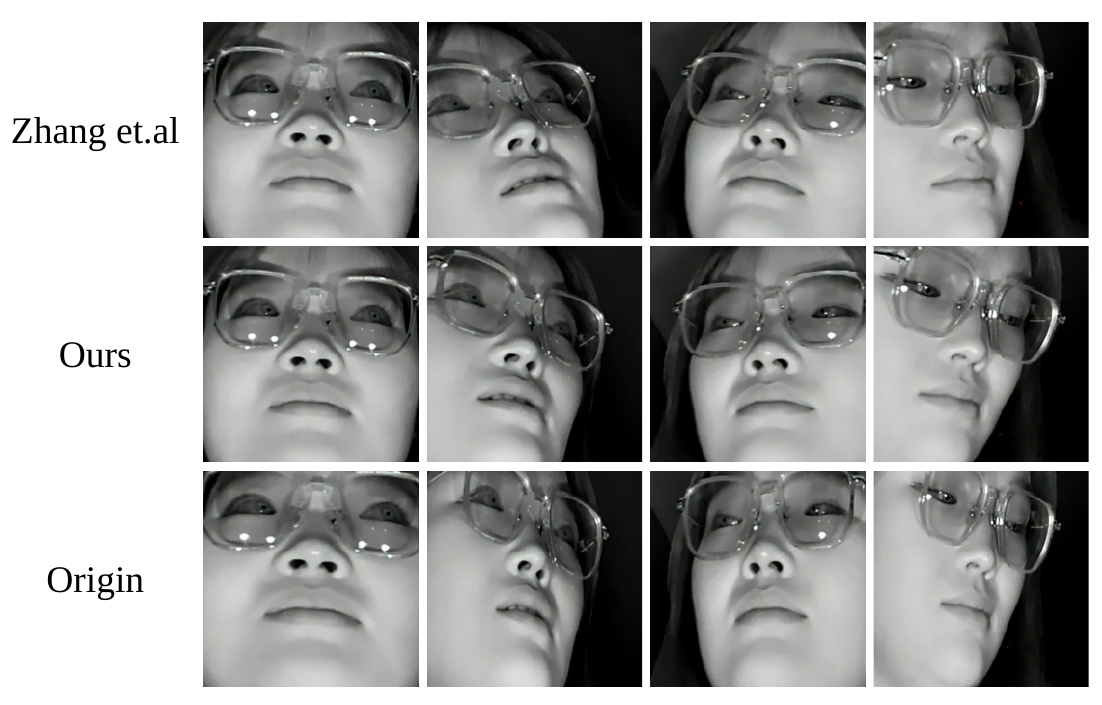}    
	\caption{We show the normalization image obtained from Zhang\etal~\cite{Zhang_2018_etra} and ours. We also visual the original images which are directly cropped from scene images. Zhang \etal rotate images based on the $x$-axis of head. It sometimes produce unstable result in extreme head pose, \eg, the second column. We modify their method and cancel such rotation. Our method has better performance which is shown in our manuscript.  \vspace{-3mm} }
	\label{fig:sup-norm}
\end{figure}

\subsection{Visualization of Normalization Images}
We show the images of different normalization methods and original images in \fref{fig:sup-norm}.
Zhang \etal~\cite{Zhang_2018_etra} rotate images based on the $x$-axis of head. It sometimes produce unstable result in extreme head pose, \eg, the second column in \fref{fig:sup-norm}. We modify their method and cancel such rotation. Our method has better performance which is shown in our manuscript.

\newpage
{
    \small
    \bibliographystyle{ieeenat_fullname}
    \bibliography{main}
}


\end{document}